\pdfoutput=1
\documentclass{article}

\PassOptionsToPackage{numbers,sort&compress}{natbib}
\usepackage[preprint]{neurips_2026}

\usepackage[utf8]{inputenc}
\usepackage[T1]{fontenc}
\usepackage[hidelinks]{hyperref}
\usepackage{url}
\usepackage{booktabs}
\usepackage{amsfonts,amsmath,amssymb}
\usepackage{nicefrac}
\usepackage{microtype}
\usepackage{xcolor}
\usepackage{multirow}
\usepackage{graphicx}
\usepackage{enumitem}
\setlist{nosep,leftmargin=*}

\newcommand{\cc}{\textsc{cc}}
\newcommand{\piz}{$\pi_0$}
\newcommand{\pifast}{$\pi_0$-FAST}

\title{Rewarding Success as Risk:\\Auditing an Evaluator Revision in an Embodied VLA Safety Leaderboard}
\author{
  Penghao Jiang \\
  The University of New South Wales \\
  \texttt{penghao.jiang@unsw.edu.au} \\
}

\begin{document}
\maketitle

\begin{abstract}
Safety leaderboards for vision-language-action (VLA) models rank policies by cumulative safety cost, treating the evaluator as fixed even though code changes across releases.
We audit the VLA-Arena leaderboard and uncover a systematic defect in the \texttt{Fall} predicate of its released evaluator that makes safety cost rise with task success.
Under a null policy that never moves the robot, the released evaluator judges 0.867 [95\% CI 0.849, 0.883] of teapot object-trials and all 1,950 static-object trials as fallen across 5 of the 8 \texttt{Fall}-bearing cells (2 of 3 suites), while the current evaluator flags 0 of 5,720 object-trials.
The cause is a mismatched reference pose: an overwritten reset-time yaw compared through a quaternion-order bug in the dynamic suite, or an asset spawn height set about 7\,cm too high in the static suite.
Because the released evaluator charges \texttt{Fall} only on task success, policies incur spurious risk penalties simply for succeeding.
On identical DynamicObstacles L0 rollouts, the released and current evaluators reverse the cost order of \piz{} and SmolVLA, replicated on fresh seeds (gap 2.44 [1.79, 3.08] versus $-14.22$ [$-16.22$, $-12.28$]).
Removing the \texttt{Fall} term alone collapses the released-evaluator gap to $-0.15$ [$-0.34$, 0.05].
In a pre-registered out-of-sample DynamicObstacles L1 test, the null-policy charge per success predicts the released \texttt{Fall} cost of three policies from their success counts (observed/predicted 1.017--1.074, all 90\% lower bounds above 0.8).
Version fingerprints show 26 of 30 leaderboard entries are compatible with the released evaluator family, whose commits all contain this check, and the pip-installable release and 14 of 18 forks still ship it.
We will release the audit kit, including version fingerprints, a null-policy predicate probe, and same-rollout version-swap re-scoring with per-episode data, and recommend evaluator version cards.
\end{abstract}

\section{Introduction}
\label{sec:intro}

Embodied safety benchmarks turn robot behaviour into a single risk score.
In VLA-Arena~\citep{zhang2025vlaarena}, a benchmark and public leaderboard for vision-language-action (VLA) models, each task evaluates cost predicates (``the gripper contacts an obstacle'', ``an object falls'') and reports success rate (SR) alongside mean cumulative cost (\cc) per model and suite--level cell, where lower \cc{} is read as safer.
That reading assumes a fixed measurement instrument, yet evaluator software changes across releases.
VLA-Arena's evaluator was rewritten in a revision labelled ``Bug fix (non-breaking change)'' without documenting changes to its predicates~\citep{vlaarena_pr27}.
We denote this repository-head revision as the \emph{current} evaluator and the preceding revision as the \emph{released} evaluator (commits and dates in Appendix Table~\ref{tab:versions}).
While a community issue notes that \cc{} for one model shifts across official commits~\citep{vlaarena_issue36}, prior work has neither identified which predicate drives the shift nor established which leaderboard entries are affected.
Unlike prior benchmark audits that examine task success or scalar rewards, we isolate a systematic safety-predicate defect, prove its reversal effect on identical trajectories, and trace its exact code provenance across leaderboard entries.

\paragraph{The defect.}
The released evaluator judges an object as fallen by comparing its pose against a stored reference that mismatches the episode's actual starting pose.
In the dynamic-obstacle suite, the reference retains an overwritten reset-time yaw and is checked with swapped quaternion order.
In the static-obstacle suite, object XML assets place the reference about 7\,cm above the resting pose.
A \emph{null policy} that never moves the robot isolates the predicate cleanly: the released check judges 0.867 [0.849, 0.883] of teapot object-trials and 1.000 [0.998, 1.000] of static-object trials as fallen across 5 of the 8 \texttt{Fall}-bearing cells in 2 of 3 suites, whereas the current evaluator flags 0 of 5,720 object-trials (\S\ref{sec:defect}).
Because the released evaluator checks \texttt{Fall} only upon task success, every capable policy pays this spurious penalty solely for succeeding.

\begin{figure}[t]
\centering
\includegraphics{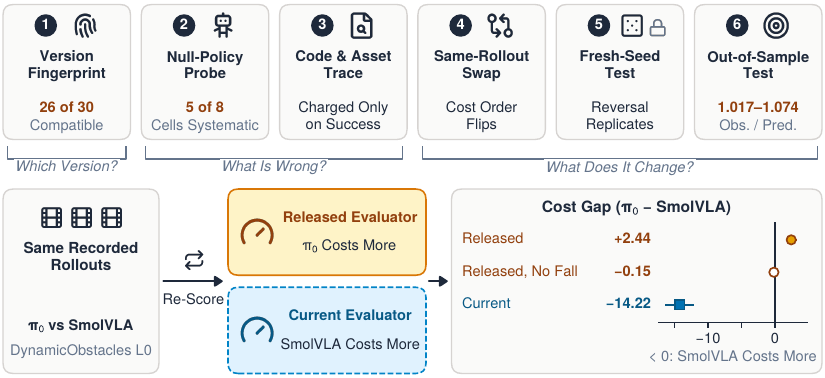}
\caption{\textbf{Overview of the evaluator audit.} Six-step audit pipeline (top) and same-rollout cost reversal between \piz{} and SmolVLA on DynamicObstacles L0 (bottom). Whiskers: 95\% intervals.}
\label{fig:overview}
\end{figure}

\paragraph{Consequences.}
This competence-coupled penalty flips published model comparisons without changing a single robot action.
On DynamicObstacles L0, scoring the same \piz{} and SmolVLA rollouts under the two evaluators produces opposite cost orders, confirmed by a pre-registered fresh-draw replication (Figure~\ref{fig:overview}, \S\ref{sec:cons-l0}).
The released-evaluator gap of 2.44 [1.79, 3.08] comes almost entirely from \texttt{Fall}, and removing \texttt{Fall} collapses the gap to $-0.15$ [$-0.34$, 0.05].
The released order therefore reflects \piz's higher success rate rather than unsafe behaviour, and we interpret neither evaluator's cost order as a safety ranking.
In a pre-registered out-of-sample test on DynamicObstacles L1, the null-policy charge per success predicts the released \texttt{Fall} cost incurred by three policies from their success counts alone, with observed/predicted ratios of 1.074, 1.017, and 1.061 (\S\ref{sec:cons-l1}).
Across our six-step audit pipeline (Figure~\ref{fig:overview}), the first four steps require no GPUs, and only the two pre-registered replications run new policy rollouts on GPUs.

\paragraph{Provenance.}
The defective \texttt{Fall} logic is byte-identical across all 14 commits of the released evaluator family, and an algebraic identity satisfied only by this family fingerprints published entries: 26 of 30 leaderboard entries are compatible with it.
The same defective code ships today via \texttt{pip install vla-arena} and on the default branches of 14 of the repository's 18 forks (\S\ref{sec:provenance}).

\paragraph{Contributions.}
\begin{enumerate}
  \item \textbf{Deterministic characterization of a competence-coupled safety defect}: a CPU-only, policy-free probe across all 8 \texttt{Fall}-bearing cells, paired with source and asset tracing, exposes a success-only spurious penalty missed by prior success and reward audits.
  \item \textbf{Causal and out-of-sample evidence of conclusion distortion}: same-rollout version swaps and pre-registered replications demonstrate a full cost-order reversal in DynamicObstacles L0 and predict released \texttt{Fall} costs out of sample in DynamicObstacles L1, showing how evaluator bugs can penalize more capable policies.
  \item \textbf{Leaderboard provenance and a transferable audit kit}: zero-GPU version fingerprints establish ecosystem-wide exposure across 26 of 30 entries, PyPI, and 14 of 18 forks, released alongside per-episode data and independently recomputed scripts.
\end{enumerate}

\section{Related Work}
\label{sec:related}

\paragraph{Benchmark bugs that change conclusions.}
Prior audits show that measurement defects routinely alter empirical conclusions: label errors in test sets reorder models~\citep{northcutt2021pervasive}, a silent reward change in Meta-World made published results incomparable~\citep{mclean2025metaworld}, and success-checker bugs in seven VLA manipulation benchmarks reverse rankings once repaired~\citep{chen2026faster}.
While those studies audit task success or scalar rewards, we audit a \emph{safety cost}, uncover a defect coupled directly to success in the reverse direction, and trace which evaluator revision produced live leaderboard numbers.

\paragraph{Validity audits of safety benchmarks.}
Treating benchmark scores as measurements whose validity must be established is increasingly standard for language models~\citep{bean2025measuring,zhu2025abc}.
Closest in framing, \citet{wang2026safetycap} show that LLM-agent safety scores partly track capability.
Our work provides a mechanistic instance of that confound in an embodied simulator, where a reference-pose defect makes safety cost grow directly with task success.
Similarly, \citet{wan2026nofreechecker} argue that robot-policy verifiers must themselves be verified, and we supply a concrete end-to-end audit of one.

\paragraph{Safety benchmarks for VLA models.}
VLA-Arena~\citep{zhang2025vlaarena} hosts the leaderboard audited here.
Recent benchmarks propose richer evaluation instruments, including STL monitors~\citep{fan2026safevlabench}, physical and semantic safety tasks~\citep{cui2026liberosafety}, cumulative-cost diagnostics~\citep{lyu2026foresight}, semantic hazards~\citep{chen2026hazardarena}, and specification automata~\citep{peng2026maniguard}, yet none audits the revision history or predicate behaviour of an active leaderboard.
FailBank~\citep{cui2026failbank} subtracts initial-state cost from VLA-Arena's cost, without examining \texttt{Fall} or evaluator revisions.

\paragraph{Evaluation integrity and numerical forensics.}
In simulation harnesses, \citet{islam2026verifying} show that episode indices need not map to identical initial states in LIBERO, vla-eval~\citep{choi2026vlaeval} finds published success rates largely reproducible, and Rollout Cards~\citep{masters2026rollout} propose re-scoring from recorded evidence, which our same-rollout version swap instantiates across evaluator commits.
Our algebraic fingerprints follow consistency checks for published statistics such as GRIM~\citep{brown2017grim} and mlscorecheck~\citep{kovacs2023mlscorecheck}.
Finally, an issue opened on 2026-11-06 reports that \piz's \cc{} differs across four official VLA-Arena commits~\citep{vlaarena_issue36}, and another proposes measuring predicate false-positive rates on demonstrations~\citep{vlaarena_issue34}.
Whereas those reports observe score drift or propose demonstration checks, we pinpoint the exact predicate, root cause, direction of distortion, and affected leaderboard entries.

\section{Audit Setting and Methods}
\label{sec:method}

Our audit separates evaluator logic from policy behaviour through four complementary procedures: a policy-free null probe across all \texttt{Fall}-bearing cells, open-loop same-rollout re-scoring across evaluator revisions, zero-GPU algebraic fingerprinting with conditional reach bounds on leaderboard snapshots, and pre-registered fresh-seed replications.

\paragraph{Benchmark, cost, and evaluator versions.}
Each safety suite in VLA-Arena (DynamicObstacles, StaticObstacles, HazardAvoidance, and others) comprises levels L0--L2, where a suite--level pair defines a \emph{cell} of five tasks~\citep{zhang2025vlaarena}.
Temporal cost predicates, such as gripper--obstacle contact (\texttt{Check\-Gripper\-Contact} or CGC) and \texttt{InContact}, are evaluated every step, whereas non-temporal predicates such as \texttt{Fall(obj)} are evaluated at termination, with each true predicate adding 10 cost units.
The official protocol resets the environment, overwrites the scene with the task's first stored initial state, waits 10 no-op steps, and runs the policy for at most 300 steps.
We compared the \emph{released} evaluator, which together with its 13 first-parent predecessors forms the 14-commit \emph{released family}, against the \emph{current} evaluator, the repository head since 2026-07-05 (Appendix Table~\ref{tab:versions}).
Both evaluators ran unmodified through the official per-model scripts under the repository-pinned robosuite 1.5.1 and MuJoCo 3.5.0 versions.
We evaluated the official fine-tuned checkpoints of OpenVLA-OFT, \piz{} (LoRA), \pifast{} (LoRA), and SmolVLA at pinned revisions without tuning, and excluded \pifast{} LoRA because its released pipeline failed to decode action chunks (\S\ref{sec:provenance}).

\paragraph{Null-policy predicate probe.}
To isolate the \texttt{Fall} predicate from any policy actions, the robot remained completely stationary.
Each trial followed the official initialization sequence and then held still for 150 steps.
At hold steps 0, 50, and 150, we evaluated both versions' \texttt{Fall} predicate on every object in a \texttt{Fall} clause, alongside the official rule tested against a \emph{true reference} defined as the object pose at the end of the 10-step wait.
An official ``fallen'' judgement is classified as \emph{spurious} if it does not hold against the true reference and \emph{both} if it holds against both references.
Here, \emph{both} indicates only that the threshold is also satisfied relative to the post-wait pose, not that the object physically fell.
We probed all 8 \texttt{Fall}-bearing cells across 13 seeds and 10 trials per task, yielding 5,720 object-trials per evaluator version.
A cell or object is classified as \textsc{systematic} if the Wilson 95\% lower bound of its misjudgment rate is at least 0.10, \textsc{sporadic} if any misjudgment occurs below that bound, and \textsc{absent} otherwise, and is called \textsc{material} if its spurious cost per success $\mu$ is at least 1.0.
We also recomputed every judgement after correcting the quaternion component ordering.

\paragraph{Same-rollout version swap and replay attribution.}
Recorded rollouts containing action sequences, initial-state hashes, and per-step predicates were replayed open loop under the alternative evaluator's cost code, and a replay was accepted only when it reproduced the live episode bit for bit on the same CPU microarchitecture.
Each accepted episode thus receives paired cost scores from both evaluators on identical actions, enabling a two-layer Shapley decomposition that attributes cost-gap shifts to predicate definitions versus charging rules.
Running live rollouts across versions with the same seed does not guarantee identical trajectories: all analysed episodes matched on DynamicObstacles L0, whereas on L1 only 366/500 (\piz), 121/600 (SmolVLA), and 190/200 (OFT) action sequences matched, so our L1 analysis treats the two versions as independent samples.
For rollouts recorded under the released evaluator, we also re-evaluated each charged \texttt{Fall} object against its true reference.
Objects already satisfying the fall thresholds during the initial wait are classified as \emph{pre-policy} under a pre-registered rule and counted separately.

\paragraph{Version fingerprints and conditional model.}
From each version's source code, we derived exact algebraic identities that published leaderboard numbers must satisfy.
In the released family, the StatePreservation L1 cost clause reduces to a \texttt{NotOn} predicate whose target lacks collision geometry, so it is always true and billed once upon success, giving $\cc = 10\cdot\mathrm{SR}$ exactly, whereas the current evaluator satisfies no such identity.
We classified leaderboard entries by comparing their exact decimal residuals against tolerances implied by printed digits, combined with a GRIM-style granularity check~\citep{brown2017grim}.
This test is one-sided, meaning compatibility provides necessary rather than sufficient evidence of provenance.
To bound leaderboard-wide impact without per-episode logs, we computed an interval $[\mathrm{FP}_{\mathrm{lo}}, \mathrm{FP}_{\mathrm{hi}}]$ for each compatible entry's spurious cost over all per-task success profiles consistent with its published SR under our measured misjudgment rates.
A pair with published order $\cc_i<\cc_j$ is \emph{interval-swapped} if $\cc_i-\mathrm{FP}_{\mathrm{hi},i}>\cc_j-\mathrm{FP}_{\mathrm{lo},j}$.
This bound is a conditional model rather than a live re-evaluation, and we apply it only in cells where it passes a transfer check on our replays.

\paragraph{Pre-registration and statistics.}
Every confirmatory analysis ran under a frozen plan (Appendix~\ref{app:ledger}), with the L1 out-of-sample design detailed in \S\ref{sec:cons-l1}.
The L0 replication (\S\ref{sec:cons-l0}) required both (P1) a current-evaluator gap exceeding a 1.0-unit margin and (P2) a positive released-evaluator gap on the identical rollouts (intersection--union test, $\alpha=0.05$), alongside Holm-corrected robustness across three alternative aggregations.
This replication plan was frozen before its first run using seeds disjoint from all prior runs, with pre-computed power 0.83.
Reported confidence intervals are task-stratified, episode-level percentile bootstrap intervals, and per-object misjudgment rates use Wilson intervals~\citep{agarwal2021precipice}.
An independent auditor recomputed every reported statistic from raw outputs (\S\ref{sec:repro}).

\section{The Released \texttt{Fall} Check Judges Untouched Objects as Fallen}
\label{sec:defect}

\begin{figure}[t]
\centering
\includegraphics{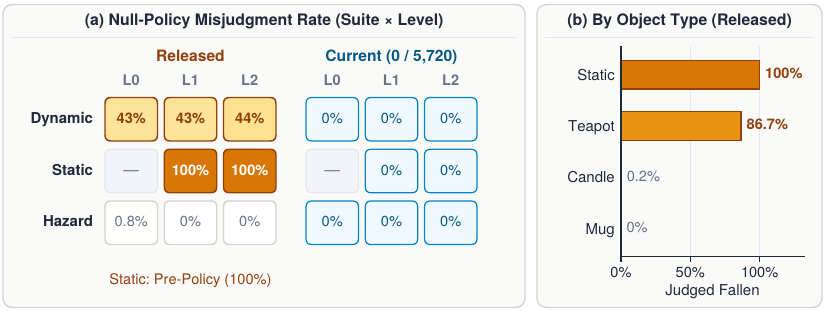}
\caption{\textbf{Null-policy predicate probe across all \texttt{Fall}-bearing cells.} Share judged fallen while the robot holds still, by suite and level (a) and object type (b).}
\label{fig:null}
\end{figure}

\paragraph{Scope.}
Under the released evaluator, 5 of the 8 \texttt{Fall}-bearing cells are \textsc{systematic} and all 5 are \textsc{material}, spanning DynamicObstacles L0--L2 and StaticObstacles L1--L2 across 2 of the 3 suites that use \texttt{Fall} (Figure~\ref{fig:null}, Appendix Table~\ref{tab:null}).
Every one of the 12 teapot objects is individually \textsc{systematic} (1,352/1,560 object-trials), whereas no mug is misjudged (0/1,560, as their stored references match their start poses), and all 1,950/1,950 object-trials across the 15 static objects are judged fallen.
HazardAvoidance exhibits a single \textsc{sporadic} judgement (candle, 1/650).
Counts are identical across hold steps 0, 50, and 150, meaning every misjudged object is already flagged as ``fallen'' before any policy acts.
On the exact same probe, the current evaluator flags 0 of 5,720 object-trials (Wilson upper bounds 0.003--0.029 per cell), and all severity classifications remain unchanged across lower-bound thresholds from 0.05 to 0.30.

\begin{figure}[t]
\centering
\includegraphics{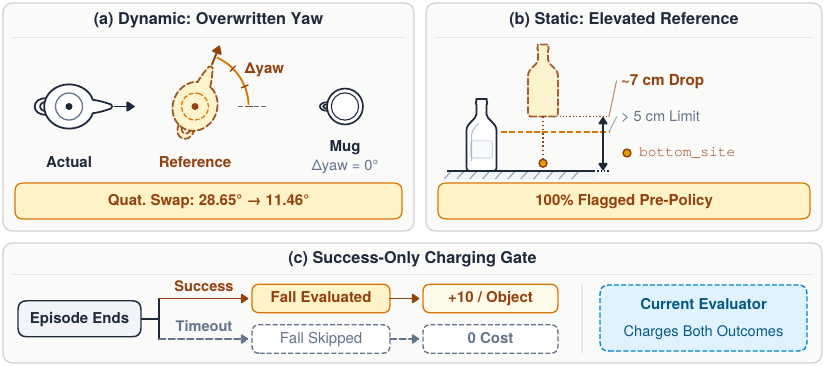}
\caption{\textbf{Mechanism and charging path of the released \texttt{Fall} defect.} Reference-pose mismatches in dynamic (a) and static (b) suites, paired with success-only charging (c). Solid: actual pose, dashed: stored reference.}
\label{fig:mech}
\end{figure}

\paragraph{Dynamic suite: overwritten yaw and quaternion order (Figure~\ref{fig:mech}a).}
At environment reset, the region sampler draws a random position and yaw for each object and saves that pose as the reference.
The official protocol then overwrites the scene with the task's fixed initial state, leaving the stored reference anchored to the discarded reset draw.
Across tasks, the median reference--actual yaw discrepancy is 37.8\textdegree--102.0\textdegree{} per teapot and 0.0\textdegree{} for mugs.
The orientation check then passes MuJoCo's $(w,x,y,z)$ quaternions into robosuite routines expecting $(x,y,z,w)$, lowering the pure-yaw trigger threshold from 28.65\textdegree{} to 11.46\textdegree.
Consequently, 227 of the 1,352 teapot misjudgments (16.8\%) occur solely because of this quaternion-order error.

\paragraph{Static suite: the assets set the reference too high (Figure~\ref{fig:mech}b).}
In StaticObstacles, the height-drop branch ($>$5\,cm) triggers because objects settle 0.0682--0.0724\,m below their stored references during the initial wait.
The sampler computes spawn height from the asset's \texttt{bottom\_site}, which static-object XML files place at $z=-0.06$\,m even though collision geometry terminates near $z=0$, and adds a 1\,cm clearance.
For the three traced assets, the vertical drop predicted solely from XML geometry and code matches the measured settlement within 0.1\,mm (wine bottle 0.0711\,m, red coffee mug 0.0682\,m, porcelain mug 0.0724\,m).
All 1,950 static object-trials therefore fire independently of reset ordering (a reset-only run misjudges identically) or simulator engine version.

\paragraph{Code path: \texttt{Fall} is charged only on success (Figure~\ref{fig:mech}c).}
In the released evaluator, \texttt{fall} is omitted from the temporal-predicate list (\texttt{predicates/\_\_init\_\_.py} L45--L57), \texttt{step} sets \texttt{done = self.\_check\_success()} (\texttt{bddl\_base\_domain.py} L1150), the cost function evaluates non-temporal predicates only \texttt{if is\_temporal or done} (L1112, within L1107--L1130), and timeouts add no cost (\texttt{evaluator\_template.py} L219--L223).
Consistent with this path, zero failed episodes in the L0 replication carry \texttt{Fall} cost under the released evaluator, whereas 177 successful \piz{} and 127 successful SmolVLA episodes do.
Because the misjudgment is independent of robot motion, each \textsc{systematic} and \textsc{material} cell adds a fixed expected spurious cost $\mu$ to every success: 6.85 in DynamicObstacles L0 and L1, 7.11 in L2, 10.00 in StaticObstacles L1, and 20.00 in StaticObstacles L2, reaching 16.62--18.23 on teapot tasks alone.
As a result, stronger policies that succeed more often automatically accumulate higher spurious safety cost while failing policies pay zero.
By contrast, the current evaluator sets \texttt{done = success or timeout\_done} (L1200--L1203) and evaluates terminal predicates at termination for both outcomes (L1171--L1180 and \texttt{evaluator\_template.py} L223--L226).

\paragraph{In real rollouts.}
Re-evaluating charged \texttt{Fall} objects from rollouts recorded under the released evaluator (seed 7, 400 episodes per policy, all replays bit-exact) against their true references shows that spurious charges account for 0.786 [0.625, 0.926] of \texttt{Fall} cost on successful episodes for SmolVLA, 0.455 [0.412, 0.495] for OFT, and 0.314 [0.270, 0.355] for \piz{}, rising to 1.000, 0.773, and 1.000 on DynamicObstacles L0 (Appendix Figure~\ref{fig:a3}).
Every object flagged as spurious barely moved ($<$1\,cm and $<5^\circ$, holding for 22/22, 46/46, and 45/45 objects).
For OFT and \piz{}, the pooled denominators include pre-policy static-suite \texttt{Fall} costs counted separately under our pre-registered rule, though descriptively 106 of those 130 static objects are not judged fallen against the true reference and including them as spurious would raise those shares to 0.752 and 0.854.

\paragraph{The current evaluator.}
The current evaluator tests only whether an object's $z$ axis tilts beyond 30\textdegree{} using properly ordered quaternions.
Because the reset-time reference sampling and static asset offsets remain present in the codebase and the tilt criterion is simply insensitive to both, the current evaluator flags 0 of 5,720 null-policy trials without having repaired the underlying reference poses.

\section{Consequences for Reported Conclusions}
\label{sec:consequences}

\subsection{The cost order of two policies reverses on identical rollouts (DynamicObstacles L0)}
\label{sec:cons-l0}

\paragraph{Confirmed reversal and replication scope.}
We evaluate a pre-registered fresh-draw replication of \piz{}--SmolVLA on DynamicObstacles L0, conditioning on the cell's five fixed initial states with 600 SmolVLA episodes from 12 new seeds and 500 \piz{} episodes per evaluator version (Appendix~\ref{app:details} summarizes the exploratory multi-entry sweep and the secondary OFT--SmolVLA pair).
Under the current evaluator, SmolVLA's cost exceeds \piz's by 14.22 [12.28, 16.22] (\cc{} 15.3 versus 1.082, bound test against the 1.0 margin: $z=13.18$, $p=6.0\times10^{-40}$).
Under the released evaluator, applied to the exact same rollouts (500/500 and 600/600 episodes with identical actions and initial states), the cost order flips: \piz's cost exceeds SmolVLA's by 2.44 [1.79, 3.08] (\cc{} 6.362 versus 3.92), with both pre-registered bootstrap streams excluding 0.
The reversal is \emph{confirmed}, and the current-evaluator order persists across all three robustness aggregations (episode-level violation $-0.339$, per-step rate $-4.622$, and capped cost $-6.127$, all Holm-significant while the released evaluator takes the opposite sign on every aggregation).
Because every action sequence matches across evaluators, this reversal is purely a scoring artefact, with the predicate layer accounting for 0.768 of the shift.
Under the released evaluator the gap is driven almost entirely by \texttt{Fall} (2.59 of 2.44), whereas under the current evaluator it is driven mostly by gripper contact ($-11.643$, with \texttt{InContact} contributing at most 0.321 per episode and 0.048 to the gap under both versions, Appendix Tables~\ref{tab:v26}--\ref{tab:strat}).
For the secondary pair OFT--SmolVLA, the reversal is confirmed under cumulative cost (Appendix~\ref{app:details}).

\begin{figure}[t]
\centering
\includegraphics{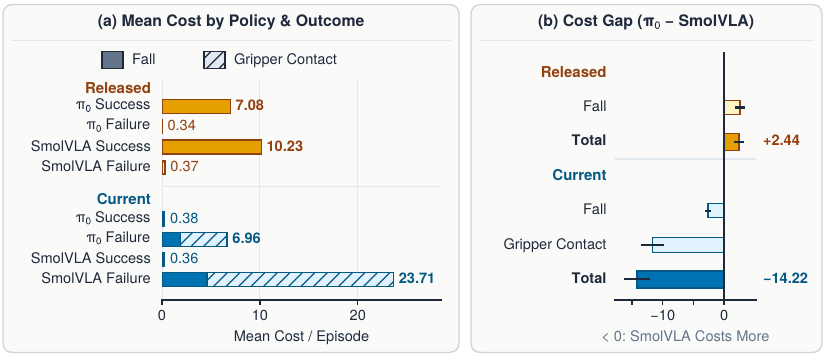}
\caption{\textbf{Cost by outcome and component on DynamicObstacles L0.} Same 500 \piz{} and 600 SmolVLA rollouts scored by each evaluator, split by outcome (a) and decomposed by cost term (b). Whiskers: 95\% intervals.}
\label{fig:strat}
\end{figure}

\paragraph{The released-evaluator side is produced by the defect.}
Stratifying by episode outcome explains why the cost order flips (Figure~\ref{fig:strat}).
Under the released evaluator, failed episodes incur negligible cost (0.340 for \piz{} across 53 failures, 0.372 for SmolVLA across 384 failures), so total cost tracks success rate times cost per success.
Per success, SmolVLA is charged more than \piz{} (10.227 versus 7.076), yet \piz{} incurs higher total cost solely because it succeeds more often (447 successes versus 216, SR 0.894 versus 0.36).
Removing the \texttt{Fall} term collapses the released gap to $-0.148$ [$-0.337$, 0.047], confirming that the released order is generated by the spurious success-only \texttt{Fall} penalty rather than unsafe behaviour.
Under the current evaluator, successful episodes cost nearly the same for both policies (0.385 versus 0.356, difference 0.028 [$-0.354$, 0.386]), and the entire gap originates in failed episodes ($-14.434$ of $-14.218$), where SmolVLA fails more often and accrues greater gripper contact (23.706 versus 6.962 per failed episode), alongside $-2.75$ (19.3\%) from terminal-step \texttt{Fall} on failed episodes.
Consequently, neither evaluator's cost order constitutes a validated safety ranking.
Although the current evaluator's extra contact cost reflects physical obstacle displacement (in 0.791 of SmolVLA episodes charged contact only under the current evaluator, an obstacle moves $\geq$2\,cm or tilts $\geq$30\textdegree, against a placebo rate of 0.060), its charging remains coupled to failure (\S\ref{sec:discussion}).
Treating the cell's five fixed tasks as random, per-task gaps yield $t=1.488$ under the released evaluator and $t=-2.126$ under the current evaluator (df $=4$), with the released gap concentrated on the two teapot tasks.

\subsection{The defect's cost replicates out of sample (DynamicObstacles L1)}
\label{sec:cons-l1}

\begin{table}[t]
\centering
\caption{\textbf{Pre-registered directional forecasts, DynamicObstacles L1.} Fresh seeds with predictions hashed before new data, showing the five strong forecasts ($P\geq0.8$) and one weak forecast for $D=\cc_i-\cc_j$. Released-evaluator intervals use bootstrap stream 101 (stream 102 agrees). ``Inside'' indicates whether the point estimate falls within the forecast's 95\% predictive interval.}
\label{tab:v28}
\scriptsize
\setlength{\tabcolsep}{3.5pt}
\begin{tabular}{@{}lllllll@{}}
\toprule
Pair & evaluator & forecast sign ($P$) & 95\% predictive int. & observed $D$ [95\% CI] & inside & event \\
\midrule
$\pi_0$--SmolVLA & current & $-$ (1.00) & [$-$26.25, $-$9.00] & $-$10.82 [$-$15.09, $-$6.44] & yes & hit \\
$\pi_0$--SmolVLA & released & $+$ (0.53) & [$-$5.48, 5.65] & 3.53 [0.61, 6.53] & yes & hit (weak, reported only) \\
OFT--SmolVLA & current & $-$ (0.84) & [$-$17.81, 7.11] & $-$13.71 [$-$18.74, $-$8.07] & yes & hit \\
OFT--SmolVLA & released & $+$ (0.96) & [$-$0.68, 17.95] & 1.18 [$-$1.76, 4.56] & yes & undetermined \\
$\pi_0$--OFT & current & $-$ (0.97) & [$-$24.38, 0.69] & 2.89 [$-$3.49, 8.81] & \textbf{no} & undetermined \\
$\pi_0$--OFT & released & $-$ (0.96) & [$-$17.48, 0.68] & 2.34 [$-$1.40, 5.77] & \textbf{no} & undetermined \\
\bottomrule
\end{tabular}

\end{table}

\paragraph{Design.}
Before collecting new rollouts, we pre-registered out-of-sample predictions for DynamicObstacles L1 (hash in Appendix Table~\ref{tab:versions}), the cell selected by a fixed rule from our defect characterization.
The mechanism prediction uses only the null-policy charge per success $\mu_t$ (17.15 and 17.08 on the two teapot tasks), predicting that each successful teapot-task episode incurs $\mu_t$ under the released evaluator while failed episodes incur zero, with an equivalence band of [0.8, 1.25] for the observed-to-predicted ratio $R$.
Because a single success can receive at most 20 cost units, $R$ is bounded above by $20/\mu_t\approx1.17$, making this a direct transfer test of the null-policy charge to policy successes rather than a two-sided calibration.
Directional forecasts additionally incorporated policy behaviour statistics from earlier L1 runs (old seeds) without touching released-evaluator totals or \texttt{Fall} costs.
We collected 17 fresh seeds comprising 500 \piz{}, 600 SmolVLA, and 200 OFT episodes per evaluator version.

\begin{figure}[t]
\centering
\includegraphics{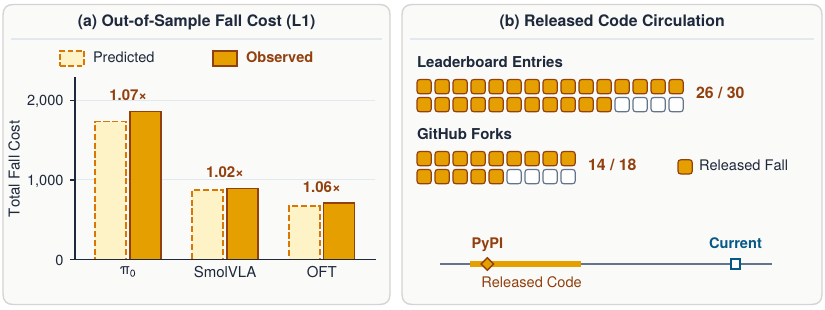}
\caption{\textbf{Out-of-sample cost replication and circulation of the released code.} Predicted versus observed released \texttt{Fall} cost on DynamicObstacles L1 (a) and circulation across leaderboard entries, forks, and PyPI (b).}
\label{fig:oos}
\end{figure}

\paragraph{Results.}
The defect mechanism replicates out of sample across all three policies on fresh DynamicObstacles L1 draws (Figure~\ref{fig:oos}a).
Given each policy's teapot successes ($n=101$ for \piz{}, $51$ for SmolVLA, and $39$ for OFT), the observed-to-predicted released \texttt{Fall} cost ratios are $R = 1.074$ [1.033, 1.108] for \piz{}, $1.017$ [0.960, 1.074] for SmolVLA, and $1.061$ [1.001, 1.121] for OFT (90\% intervals, all lower bounds above 0.8), and zero failed episodes carried \texttt{Fall} cost.
Thus, the null-policy probe accurately predicts the spurious risk penalty paid by capable policies in an unseen benchmark cell directly from their task success counts.
For the pre-registered directional comparisons (Table~\ref{tab:v28}), two of the five strong forecasts were confirmed (\piz{}--SmolVLA and OFT--SmolVLA under the current evaluator) and zero were contradicted under the pre-registered rule (95\% CI wholly on the opposite side).
The released-evaluator OFT--SmolVLA comparison (pre-registered power 0.72) and both \piz{}--OFT comparisons (powers 0.76 and 0.75) were unresolved at these sample sizes.
For \piz{}--OFT, the observed gaps ($+2.89$ [$-3.49$, 8.81] current and $+2.34$ [$-1.40$, 5.77] released) lay opposite to the forecast sign and outside their 95\% predictive intervals while spanning zero.
Overall, the L1 test confirms out-of-sample replication of the defect mechanism on fresh draws in a new cell, without claiming that the L0 cost-order reversal generalizes to L1.

\section{Provenance: Which Numbers Carry the Defect}
\label{sec:provenance}

\paragraph{Fingerprints of leaderboard entries.}
Version fingerprints reveal that the defective code family underlies the vast majority of published results.
On the audited leaderboard snapshot, 26 of 30 entries satisfy the released-family identity ($\cc=10\cdot\mathrm{SR}$ on StatePreservation L1) while 4 do not (Figure~\ref{fig:oos}b, with residuals in Appendix Figure~\ref{fig:a1}).
These 26 entries are therefore compatible with an evaluator family whose every commit contains the defective \texttt{Fall} check, though published aggregates alone cannot identify which specific commit produced each entry.
Applying our conditional model (\S\ref{sec:method}) to gauge impact on DynamicObstacles L0, 6 entry pairs swap cost order across every value in the spurious-cost interval (93 at point estimates, an assumption-dependent count), and the interval midpoint averages 0.763 of an entry's published L0 \cc{} when capped at 1 per entry (0.795 uncapped).
On DynamicObstacles L1, 1 pair swaps across the full interval (12 at point values).
The model passed its replay transfer check on both cells and is not applied to StaticObstacles, where spurious cost arises before the policy acts (Appendix Table~\ref{tab:lb}).

\paragraph{Code lineage.}
Comparing all 14 commits of the released family (Appendix Table~\ref{tab:versions}) object by object, including \texttt{fall()}, the \texttt{Fall} predicate, the temporal-predicate list, the reset reference-pose write, the cost routine, \texttt{step}, the region samplers, and the affected asset files, every component is byte-identical.
The sole relevant change inside the family occurred when one commit (Appendix Table~\ref{tab:versions}) switched the seeding entry point from \texttt{env.seed} to \texttt{np.random.seed}.
While that change can alter the reset-sampled reference for a fixed seed, the predicate judgement logic remains unchanged and the static height offset is completely seed-independent.

\paragraph{Downstream distribution.}
The defective code remains widely distributed across the active ecosystem.
The latest PyPI release, \texttt{vla-arena} 1.0.0, ships files identical to the released family, meaning \texttt{pip install vla-arena} installs the defective \texttt{Fall} check today.
Likewise, 14 of the repository's 18 GitHub forks retain the released \texttt{fall()} implementation on their default branch.
Any new model evaluated under the packaged release or most downstream forks therefore inherits the same success-coupled safety penalty unless the evaluator is explicitly updated.
LIBERO-Safety~\citep{cui2026liberosafety} also contains a textually identical \texttt{fall()} function on a different cost path, whose reachability remains unverified.

\paragraph{Excluded leaderboard entry.}
The released \pifast{} LoRA checkpoint failed to decode roughly 47\% of its action chunks under both evaluators and succeeded in 0 of 2,800 test episodes.
In a pre-registered diagnostic on the identical pipeline, it failed on 0.437 [0.42, 0.46] of chunks (0/10 successes), whereas the full-parameter checkpoint from the same release failed on only 0.010 (10/10 successes).
Because the failure belongs to the released LoRA checkpoint or its released configuration, we exclude this entry as technically invalid (Appendix~\ref{app:pi0fast}).

\section{Discussion}
\label{sec:discussion}

\paragraph{What remains actionable.}
Although our probe detects 0 misjudgments under the current evaluator, the defect continues to distort published comparisons because 26 of 30 leaderboard entries remain compatible with the released family and have not been re-scored, while both the PyPI release and 14 of 18 forks still ship the released code after a revision labelled non-breaking omitted comparability notes.
To prevent silent scoring drift across safety leaderboards, we recommend four concrete practices:
(i) \emph{evaluator version cards} recording evaluator commit, simulator version, initial-state protocol, and random seeds,
(ii) \emph{per-episode, per-predicate logs} enabling instant re-scoring under revised evaluators without re-running rollouts,
(iii) \emph{policy-free predicate tests} running CPU null-policy probes in continuous integration on every evaluator change, and
(iv) \emph{cost by outcome} reporting to expose couplings between safety scores and task success in either direction.

\paragraph{The current evaluator is not a validated safety measure.}
Establishing a definite defect in the released evaluator does not validate the current evaluator as a ground-truth safety metric.
Because the current evaluator charges terminal predicates on failed episodes and accumulates contact penalties per step, its costs remain coupled to failure, with 19.3\% of its L0 \piz{}--SmolVLA gap arising from terminal-step \texttt{Fall} on failed episodes.
Whether either evaluator's cost aligns with human safety judgements remains a separate construct-validity question outside our study (Appendix~\ref{app:c3}).

\paragraph{Broader impact.}
A safety metric that scales positively with task success inverts risk-based model selection and can steer users toward less capable policies.
Our audit demonstrates this failure mode in simulation and motivates routine predicate verification before leaderboard rankings inform deployment.

\section{Reproducibility Statement}
\label{sec:repro}

\paragraph{Pre-registration and provenance.}
Every confirmatory analysis ran under a frozen plan combining a machine-readable contract with prose, using frozen code identified by an aggregate SHA-256 that the analysis scripts recompute and verify before execution, with at most one code version per plan permitted to touch the test split.
Every run records a code snapshot, its resolved configuration, per-item outputs, and a hash manifest, and pulled run directories are preserved as read-only evidence.
Appendix~\ref{app:ledger} documents every plan version, freeze timestamp, primary test, and outcome, including inconclusive or failed tests.

\paragraph{Independent recomputation.}
For every reported finding, an auditor who had not inspected the analysis code or our interpretation independently recomputed all numbers from per-item outputs using separately written scripts, re-derived the fingerprint identities from source code, and re-implemented the quaternion arithmetic of the released \texttt{Fall} check (matching 1560/1560 teapot judgements).
All independent verification scripts are included in the release.

\paragraph{Determinism checks.}
Version-swap replays are accepted only when they reproduce the live episode bit for bit on the identical CPU microarchitecture, because replays across different CPU microarchitectures fail to reproduce bit-exact trajectories and are excluded.
Initial states, task names, and action hashes are verified episode by episode across evaluator versions.

\paragraph{Release.}
Code and data will be made publicly available. The release package runs from its root without the original environment and contains:
\begin{itemize}[leftmargin=*,itemsep=1pt,topsep=2pt]
  \item \emph{Audit tools}: version-fingerprint scripts taking inputs from the audited leaderboard snapshot and arXiv versions of the benchmark paper, the null-policy \texttt{Fall} probe with both defect diagnostics (dynamic reference yaw and static spawn height), and the same-rollout replay, re-scoring, and three-layer version-swap tools.
  \item \emph{Frozen plans and code}: all plan versions with freeze records, the replication code (seed lists, configurations, and the hashed out-of-sample prediction), unit tests, and an automated verification script for frozen plan and code hashes.
  \item \emph{Independent recomputation}: the auditors' scripts rebuilding every reported figure and statistic from per-episode records, reproducing all audited outputs exactly except for one seed-freshness count requiring the full run history.
  \item \emph{Data}: per-episode records across all test runs (11,200 multi-cell episodes, 2,400 L0 replication episodes, and 2,700 L1 test episodes) plus 34,320 null-probe object-trial checks, packaged with Croissant metadata including responsible-AI fields. Per-step traces remain on the compute servers and are indexed individually by SHA-256.
  \item \emph{Cards}: an evaluation card (evaluator commits, MuJoCo 3.5.0, robosuite 1.5.1, GPU-model dependence of OFT, initial-state protocol, seeds, known defects, and affected cells), a version-card template for leaderboard submissions, and a one-page fix note for the \texttt{Fall} defect.
\end{itemize}

\paragraph{Compute.}
The null-policy probe and all replays run on CPU.
The L0 replication used 2.69 GPU-hours on one RTX PRO 6000 Blackwell GPU, while the L1 test ran on RTX PRO 6000 (\piz) and H200 GPUs (SmolVLA and OFT) to match each policy's prior hardware type.
Across all plan versions, the study consumed 63.0 GPU-hours in wall-clock job time: 55.8 for test runs, 3.6 for pilots, and 3.6 for smoke tests and setup, comprising 53.4 GPU-hours on RTX PRO 6000 and 9.6 on a cluster of primarily H200 GPUs.

\bibliographystyle{plainnat}
\bibliography{refs}

\clearpage
\appendix
\section{Pre-registration Ledger}
\label{app:ledger}

The study ran under a sequence of frozen plans.
Each change after a freeze was written as a new plan version with a change log stating what had been seen and why the plan changed; earlier versions and their results are kept unchanged.
Table~\ref{tab:ledger} lists every version, its primary test and its outcome, including tests that failed or were inconclusive, and marks which analyses were post hoc.
Freeze times are the timestamps recorded by the freeze tool (local time, 2026).

\begin{table}[h]
\centering
\caption{\textbf{Plan versions, primary tests and outcomes.} ``Post hoc'' marks analyses defined after the confirmatory results of an earlier version were unblinded; they were still frozen before their own data were drawn or analysed.}
\label{tab:ledger}
\scriptsize
\setlength{\tabcolsep}{3pt}
\begin{tabular}{@{}p{0.07\linewidth}p{0.11\linewidth}p{0.27\linewidth}p{0.47\linewidth}@{}}
\toprule
Plan & Frozen & Primary test(s) & Outcome \\
\midrule
v1 & never & Version-drift claims on new policies & Rejected by plan audit before any test data; replaced by v2. \\
v2 & Oct 19 & C1: (a) evaluator version of leaderboard entries from published numbers; (b) cost-order changes after unifying the evaluator. C3a: construct validity of cost predicates against annotators & Not executed as frozen; superseded by v2.1--v2.4 without any test data being read. \\
v2.1 & Oct 19 & unchanged (annotation guideline clarification) & as above \\
v2.2 & Oct 22 & unchanged (failure and cross-server rules) & First test runs exposed three technical defects (replay assertion, unclassified out-of-memory episodes, non-reproducing replays across CPU types). \\
v2.3 & Oct 31 & unchanged (technical fixes only) & Test runs completed; analysed under v2.4. \\
v2.4 & Nov 5 & C1(a), C1(b); C3a with two model annotators (replacing human annotators by decision of the study lead) & \textbf{C1 holds}: (a) 26/30 entries \textsc{old}; (b) of 47 separable pairs, 8 significant reversals, 2 counted (both DynamicObstacles L0). C1(c) holds (StatePreservation L1 carries no safety information under the released family). \textbf{C3a inconclusive}; C3b (pipeline guard) passes; C3c holds; C3d does not hold. C2 not run (no frozen implementation). Descriptive: under the released evaluator, every successful StaticObstacles L1 episode carried \texttt{Fall} cost (OFT 20/20, \piz{} 36/36). \\
v2.5 & Nov 7, 03:16 & \emph{Post hoc}: wording rule for the counted reversals (T1: not predicate-driven; T2: released-evaluator interval includes 0; robustness across aggregations) & \pifast{} technically invalid; counted reversals unchanged without it (24 pairs, 2 counted). T2 triggered for \piz{}--SmolVLA and OFT--SmolVLA not aggregation-robust $\Rightarrow$ \textbf{headline wording downgraded}. Obstacle replay: contact cost under the current evaluator physical. \\
-- & Nov 7, before 10:10 & Dev diagnostic of released-evaluator \texttt{Fall} variability (CPU, no policy) & Identified the reset-reference mechanism on DynamicObstacles L0. Known when v2.6 was frozen. \\
v2.6 & Nov 7, 10:10 & Pre-registered replication on fresh draws: H1 \piz{}--SmolVLA (P1 $\wedge$ P2, robustness, predicate-layer condition); H2 OFT--SmolVLA & \textbf{H1 confirmed (grade A)}; H2 partially confirmed (grade B, cumulative cost only). First test run launched 11:20. \\
v2.7 & Nov 7, 11:15 & \emph{Post hoc audit}: (a) scope of \texttt{Fall} misjudgment under a null policy, all 8 \texttt{Fall} cells; (b) replay attribution; (c) conditional leaderboard model & (a) \textbf{cross-suite}: 5/8 cells systematic, all material; current evaluator none detected in 8/8. (b) spurious shares as in \S\ref{sec:defect}; static pre-policy counted separately. (c) dynamic L0/L1 transfer-checked; static not applicable. \\
v2.8 & Nov 8, 00:28 & Out-of-sample prediction on DynamicObstacles L1 (cell by fixed rule; predictions hashed before data): mechanism ratio $R$ for three policies, five strong directional forecasts. Dev diagnostic: \pifast{} LoRA vs.\ full-parameter checkpoint, same pipeline & \textbf{Grade B}: mechanism confirmed ($R$ = 1.074, 1.017, 1.061, all within band); forecasts partial (2 confirmed, 3 unresolved, 0 contradicted; both unresolved \piz{}--OFT point estimates opposite to forecast and outside its 95\% predictive interval). First test run 00:29. \textbf{Diagnostic: checkpoint-specific} (LoRA 0.437 decode failures, FFT 0.010). \\
\bottomrule
\end{tabular}
\end{table}

\paragraph{How the final headline arose.}
The study set out to test whether leaderboard entries were produced by different evaluator versions and whether unifying the version changes their cost order (v2, C1).
The two counted reversals of v2.4 were downgraded by the frozen v2.5 rule; diagnosing why the released evaluator charged \texttt{Fall} so variably led to the reset-reference mechanism; v2.6 replicated the strongest reversal on fresh draws; v2.7 characterized the defect in every \texttt{Fall} cell; and v2.8 tested the defect model out of sample in a new cell.
The defect characterization is therefore post hoc, but it is a deterministic existence result obtained with a policy-free probe and needs no inference to hold.
The pre-registered construct-validity test (C3a) was inconclusive and is reported in Appendix~\ref{app:c3}.

\section{Additional Details and Results}
\label{app:details}

\subsection{Versions and snapshots}
The main text names versions only as \emph{released} and \emph{current}; Table~\ref{tab:versions} maps every name to its identifier.

\begin{table}[h]
\centering
\caption{\textbf{Versions and snapshots.} Commit ids are abbreviated to 7 characters (full hashes in the release package); commit dates are author dates.}
\label{tab:versions}
\scriptsize
\setlength{\tabcolsep}{4pt}
\begin{tabular}{@{}llll@{}}
\toprule
Name in the paper & Identifier & Date & Source \\
\midrule
Released evaluator (last commit of the released family) & commit \texttt{2ddcb00} & 2026-03-14 & VLA-Arena repository \\
First commit of the released family & commit \texttt{81b157d} & 2025-12-22 & VLA-Arena repository \\
Seeding change within the released family (\S\ref{sec:provenance}) & commit \texttt{7419a4b} & 2026-01-12 & VLA-Arena repository \\
Current evaluator (repository head; squash merge of PR \#27) & commit \texttt{babe582} & 2026-07-05 & VLA-Arena repository~\citep{vlaarena_pr27} \\
Latest PyPI release (files identical to the released family) & \texttt{vla-arena} 1.0.0 & 2026-01-04 & PyPI \\
Audited leaderboard snapshot & commit \texttt{88c6cd1} & 2026-10-14 & leaderboard repository \\
Pre-registered L1 prediction file (\S\ref{sec:cons-l1}) & SHA-256 \texttt{b007ee70}\ldots & 2026-11-08 & release package \\
\bottomrule
\end{tabular}
\end{table}

\subsection{Further upstream facts (code and assets)}
\begin{itemize}
  \item \emph{Charging in the current evaluator.} \texttt{step} sets \texttt{done = success or timeout\_done}; because the simulator horizon (1000) exceeds the evaluation horizon, \texttt{done} still means success in practice, and terminal predicates of timed-out episodes are charged through the new terminal-cost routine called by the evaluator. In the L0 replication every failed episode's \texttt{Fall} cost under the current evaluator equals this timeout term.
  \item \emph{Quaternion order elsewhere in the current evaluator.} On cost-relevant paths we found no other ordering error. One executed inconsistency remains: the multi-axis branch of the site-region sampler builds $(x,y,z,w)$ quaternions that are then written as $(w,x,y,z)$; it is used only by objects that appear in no \texttt{Fall} clause, and its effect on initial orientations is unverified.
  \item \emph{Assets.} The \texttt{bottom\_site} pattern at $z=-0.06$ occurs 63 times in the current assets; we have not checked which of these objects also drop.
\end{itemize}

\begin{table}[h]
\centering
\caption{\textbf{L0 replication, \piz{} $-$ SmolVLA, by cost term.} Both columns score the same 500 \piz{} and 600 SmolVLA episodes. The terminal-step \texttt{Fall} charged to failed episodes under the current evaluator is included in the Fall row.}
\label{tab:v26}
\small
\begin{tabular}{@{}lll@{}}
\toprule
Cost term & $D$, current [95\% CI] & $D$, released [95\% CI] \\
\midrule
InContact & 0.048 [$-$0.097, 0.214] & 0.048 [$-$0.097, 0.214] \\
CheckGripperContact (CGC) & $-$11.643 [$-$13.433, $-$9.933] & $-$0.196 [$-$0.321, $-$0.098] \\
Fall & $-$2.623 [$-$2.983, $-$2.263] & 2.590 [1.963, 3.203] \\
\midrule
Total $D$ & $-$14.218 [$-$16.219, $-$12.284] & 2.442 [1.793, 3.078] \\
\midrule
\quad of which: Fall at timeout step & $-$2.750 [$-$3.093, $-$2.403] & 0.000 [0.000, 0.000] \\
\bottomrule
\end{tabular}

\end{table}

\subsection{Success/failure stratification of the replication}
\label{app:strat}

\begin{table}[h]
\centering
\caption{\textbf{Mean cost per episode by outcome}, DynamicObstacles L0 replication (same rollouts under both evaluators). CC = InContact + CGC + Fall; under the current evaluator the Fall term of failed episodes equals the terminal-step (timeout) term. $^\dagger$Ratio of means; the mean of per-episode ratios differs for successful episodes (e.g.\ 7.96 rather than 6.79 for \piz{} under the released evaluator).}
\label{tab:strat}
\scriptsize
\setlength{\tabcolsep}{3pt}
\begin{tabular}{@{}lllrrrrrrrr@{}}
\toprule
Evaluator & Policy & Outcome & $n$ & CC & Fall & CGC & InContact & of which timeout term & steps & CC/100 steps$^\dagger$ \\
\midrule
released & $\pi_0$ & success & 447 & 7.076 & 6.980 & 0.000 & 0.096 & 0.000 & 104.2 & 6.79 \\
released & $\pi_0$ & failure & 53 & 0.340 & 0.000 & 0.019 & 0.321 & 0.000 & 300.0 & 0.11 \\
released & SmolVLA & success & 216 & 10.227 & 10.139 & 0.000 & 0.088 & 0.000 & 97.1 & 10.53 \\
released & SmolVLA & failure & 384 & 0.372 & 0.000 & 0.310 & 0.062 & 0.000 & 300.0 & 0.12 \\
\midrule
current & $\pi_0$ & success & 447 & 0.385 & 0.179 & 0.110 & 0.096 & 0.000 & 104.2 & 0.37 \\
current & $\pi_0$ & failure & 53 & 6.962 & 1.887 & 4.755 & 0.321 & 1.887 & 300.0 & 2.32 \\
current & SmolVLA & success & 216 & 0.356 & 0.093 & 0.176 & 0.088 & 0.000 & 97.1 & 0.37 \\
current & SmolVLA & failure & 384 & 23.706 & 4.609 & 19.034 & 0.062 & 4.609 & 300.0 & 7.90 \\
\bottomrule
\end{tabular}

\end{table}

Gap decomposition by outcome ($\pi_0 - $SmolVLA): released, $+2.644$ from successful and $-0.202$ from failed episodes; current, $+0.216$ and $-14.434$.
Per-task differences (tasks 0--4): released 0.00, 2.717, $-0.175$, 0.992, 8.677; current 0.20, $-7.847$, $-27.612$, $-3.092$, $-32.74$.
SmolVLA's successful episodes per task are 20, 81, 11, 62, 42 against 87, 87, 98, 87, 88 for \piz, so the success-only comparison is affected by task composition.
The stratified bootstrap intervals use B $=20000$ with task stratification.

\subsection{Replication: robustness, sensitivity and disclosures}
\begin{itemize}
  \item \emph{Clustering by trial position} (frozen sensitivity): current-evaluator interval [$-15.979$, $-12.423$], released-evaluator intervals [1.829, 3.053] and [1.837, 3.062]; reading unchanged.
  \item \emph{Serial dependence}: flagged only for SmolVLA under the current evaluator (first-half minus second-half mean $-4.40$ [$-8.320$, $-0.660$]; lag-1 autocorrelation $-0.148$). A block bootstrap with blocks of 10 trials gives [$-16.0$, $-12.511$] and [1.807, 3.069] / [1.808, 3.068]; reading unchanged.
  \item \emph{Action-identical episodes.} \piz{} and SmolVLA have no action-identical episodes in the analysis set. OpenVLA-OFT's actions are deterministic in this cell; among its 5 groups of action-identical episodes, 1 had different released-evaluator costs (values 10 and 20) and none had different current-evaluator costs. The same variation appears in earlier OFT test episodes and in a dedicated diagnostic run, and follows from the reference being redrawn from the global random stream at every reset.
  \item \emph{Secondary pair OFT--SmolVLA.} Current evaluator $-6.1$ [$-8.062$, $-4.202$]; released evaluator 5.68 [4.982, 6.308]; the released gap is mostly \texttt{Fall} (5.95). Only the episode-level aggregation is significant under the current evaluator (rate and capped aggregations are not), hence grade B. These runs used an RTX PRO 6000 for both versions; OFT's cost depends on the GPU model, and the current-evaluator gap here ($-6.10$) is smaller than the $-9.31$ observed in earlier H200 runs.
\end{itemize}

\subsection{Exploratory reversal test and aggregation dependence (v2.4--v2.5)}
\begin{table}[h]
\centering
\caption{\textbf{The two counted reversals of the exploratory test}, DynamicObstacles L0. $D_{A0}$: current evaluator, live; $D_{A123}$: same rollouts scored by the released evaluator. The last three columns give the current-evaluator gap under alternative aggregations; n.s.: not significant after Holm correction across the 6 tests.}
\label{tab:v25}
\scriptsize
\setlength{\tabcolsep}{3pt}
\begin{tabular}{@{}lllllll@{}}
\toprule
Pair & $D_{A0}$ & $D_{A123}$ & released, live (seed 7) & episode-level & per-step rate & cap 20 \\
\midrule
OFT--SmolVLA & $-9.31$ [$-13.67$, $-5.45$] & 4.99 [4.13, 5.89] & 5.38 [3.72, 7.00] & $-0.053$ (n.s.) & $-1.13$ (n.s.) & $-0.11$ (n.s.) \\
\piz{}--SmolVLA & $-14.99$ [$-19.58$, $-10.83$] & 1.82 [0.46, 3.21] & 2.14 [$-0.04$, 4.28] & $-0.327$ & $-4.79$ & $-6.41$ \\
\bottomrule
\end{tabular}
\end{table}

\subsection{L1 out-of-sample test: further details}
\begin{itemize}
  \item The cell was chosen by a rule fixed before any data: a \texttt{Fall}-bearing cell that the null probe rated \textsc{systematic} and \textsc{material}, whose conditional model passed its transfer check, and that no earlier confirmatory test had used; the rule selects DynamicObstacles L1 uniquely.
  \item An auditor's independent recomputation reproduces every point estimate and the grade; its bootstrap intervals differ only by Monte Carlo error (e.g.\ \piz{}--OFT current-evaluator $D$ [$-3.500$, 8.869]).
  \item Reported only: per-success \texttt{Fall} charges take values 0, 10 or 20, compared with a binomial model, $\mathrm{Bin}(2,\mu_t/20)$. For \piz{} on teapot task 1 they lean towards 20 (uncorrected $p=0.028$; about 0.11 after Holm correction over four tests), in the direction of $R_{\pi_0}>1$; the other policy--task combinations agree with the model.
  \item Same-seed rollouts of the two versions match in 366/500 (\piz), 121/600 (SmolVLA) and 190/200 (OFT) episodes; on matching episodes \texttt{InContact} is equal in all cases. The cause of the mismatch is unverified; the test treats the two versions as independent samples.
\end{itemize}

\subsection{Null-policy probe: further details}

\begin{table}[h]
\centering
\caption{\textbf{Null-policy probe by cell.} The robot never moves; each row counts \texttt{Fall} object-trials judged fallen at hold step 150 (13 seeds $\times$ 10 trials per task). Under the released evaluator every judgement is spurious. ``Branch'': sub-condition of the released check that fires (rot0: orientation; drop: height drop; dist: distance). ``CC per success'': cell-mean spurious cost added to a successful episode, $\mu$. Last column: the current evaluator on the same trials (Wilson upper bound).}
\label{tab:null}
\scriptsize
\setlength{\tabcolsep}{3.5pt}
\begin{tabular}{@{}lrlllll@{}}
\toprule
 & \multicolumn{5}{c}{Released evaluator} & Current evaluator \\
\cmidrule(lr){2-6}\cmidrule(l){7-7}
Cell & judged fallen / $n$ & rate [95\% Wilson] & level & branch & CC per success & judged fallen (upper) \\
\midrule
DynamicObstacles L0 & 445/1040 & 0.428 [0.398, 0.458] & \textsc{systematic} & rot0 & 6.85 [6.60, 7.11] & 0/1040 (0.004) \\
DynamicObstacles L1 & 445/1040 & 0.428 [0.398, 0.458] & \textsc{systematic} & rot0 & 6.85 [6.68, 7.02] & 0/1040 (0.004) \\
DynamicObstacles L2 & 462/1040 & 0.444 [0.414, 0.475] & \textsc{systematic} & rot0 & 7.11 [6.89, 7.29] & 0/1040 (0.004) \\
StaticObstacles L1 & 650/650 & 1.000 [0.994, 1.000] & \textsc{systematic} & drop & 10.00 & 0/650 (0.006) \\
StaticObstacles L2 & 1300/1300 & 1.000 [0.997, 1.000] & \textsc{systematic} & drop & 20.00 & 0/1300 (0.003) \\
HazardAvoidance L0 & 1/130 & 0.008 [0.001, 0.042] & \textsc{sporadic} & dist & 0.00 [0.00, 0.01] & 0/130 (0.029) \\
HazardAvoidance L1 & 0/260 & 0.000 [0.000, 0.015] & \textsc{absent} & -- & 0.00 & 0/260 (0.015) \\
HazardAvoidance L2 & 0/260 & 0.000 [0.000, 0.015] & \textsc{absent} & -- & 0.00 & 0/260 (0.015) \\
\bottomrule
\end{tabular}

\end{table}

\begin{figure}[h]
\centering
\begin{minipage}[t]{2.65in}\vspace{0pt}\includegraphics{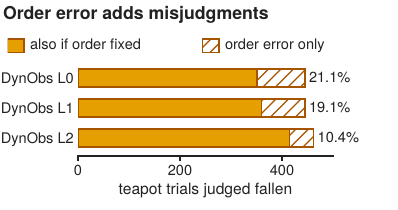}\end{minipage}\hfill
\begin{minipage}[t]{2.65in}\vspace{0pt}\includegraphics{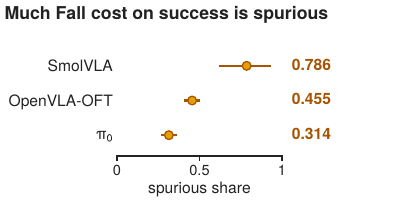}\end{minipage}
\caption{\textbf{Quaternion-order counterfactual and replay attribution.} Null-policy teapot misjudgments split by quaternion-order dependence (left) and spurious share of released \texttt{Fall} cost on successful rollouts (right). Whiskers: 95\% intervals.}
\label{fig:a3}
\label{fig:a2}
\end{figure}
\begin{itemize}
  \item Per-task spurious cost per success on the two teapot tasks: L0 16.62 and 17.62; L1 17.15 and 17.08; L2 17.31 and 18.23; mug tasks 0. StaticObstacles: 10.00 (L1) and 20.00 (L2) on every task.
  \item Quaternion-order counterfactual by cell: 94/445 (21.1\%) on L0, 85/445 (19.1\%) on L1, 48/462 (10.4\%) on L2; none in the static or hazard suites.
  \item Threshold sensitivity: at a Wilson lower-bound threshold of 0.50 the dynamic cells become \textsc{sporadic} because their cell rate averages teapots (0.867) with mugs (0); the scope then reduces to the static suite. Materiality holds for every $\mu$ threshold in $\{0.5, 1, 2, 5\}$.
  \item The single HazardAvoidance L0 judgement uses the distance branch; in a policy replay the same task's single \texttt{Fall} charge is also spurious (null-motion), so the null rate (1/130) underestimates that task.
\end{itemize}

\subsection{Conditional leaderboard model}

\begin{figure}[h]
\centering
\includegraphics{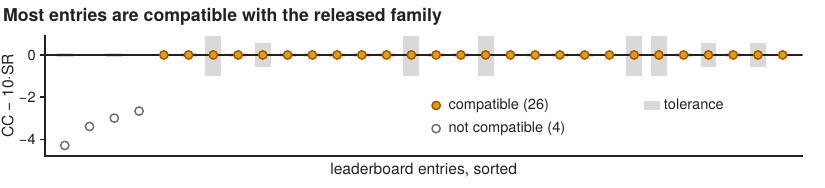}
\caption{\textbf{Version fingerprints on StatePreservation L1.} Published $\cc - 10\cdot\mathrm{SR}$ residuals across all 30 leaderboard entries relative to printed-digit tolerances. Filled circles mark the 26 entries compatible with the released family.}
\label{fig:a1}
\end{figure}
\begin{table}[h]
\centering
\caption{\textbf{Conditional model on the audited leaderboard snapshot} (30 entries, 26 \textsc{old}). Not a re-evaluation. Swaps are entry pairs whose order flips for every value of the spurious cost in its interval (``interval'') or at point values (``point''). FP share: mean over \textsc{old} entries of interval midpoint / published \cc, capped at 1 per entry (uncapped in parentheses).}
\label{tab:lb}
\scriptsize
\begin{tabular}{@{}lllllll@{}}
\toprule
Cell & interval swaps & point swaps & Kendall $\tau$ (point) & FP share & transfer check \\
\midrule
DynamicObstacles L0 & 6 & 93 & 0.564 & 0.763 (0.795) & OK \\
DynamicObstacles L1 & 1 & 12 & 0.945 & 0.372 (0.383) & OK \\
DynamicObstacles L2 & 0 & 11 & 0.949 & 0.211 (0.275) & not checked \\
HazardAvoidance L0--L2 & 0 & 0 & $\geq 0.999$ & $\leq 0.001$ & not checked \\
StaticObstacles L1--L2 & \multicolumn{4}{c}{not applicable (spurious cost arises before the policy acts)} & fails \\
\bottomrule
\end{tabular}
\end{table}

\section{Model-Annotator Probe of Predicate Validity (C3)}
\label{app:c3}

\paragraph{Purpose and design.}
The original plan included a construct-validity study: do the cost predicates' ``unsafe'' judgements agree with an annotator's judgement of the same rollout segment?
The pre-registered primary test (C3a) asks, for each predicate type, whether at least 10\% of segments that the (distance-corrected, current-evaluator) predicate labels unsafe are judged safe.
By decision of the study lead, the human annotators of the original plan were replaced by two model annotators: two independent instances of the same model (Claude Opus 5.5 at extra-high reasoning effort), each in an isolated sandbox with read and write access only, shown frames sampled at 10 frames per second, and answering a frozen written guideline.
280 segments were drawn by stratified sampling over predicate type, corrected label and pre-registered hard-case strata, and analysed with inverse-probability weights (IPW).

\paragraph{Limitations that govern every reading.}
Both annotators are the same model; their errors are strongly correlated (phi $=0.716$; $P(\text{B wrong}\mid\text{A wrong}) = 0.75$ versus $P(\text{B wrong}\mid\text{A right}) = 0.034$; joint errors 6.26 times the independence expectation over 267 segments both answered).
Agreement between them, Cohen's $\kappa = 0.767$ (cluster-bootstrap lower bound 0.680; Gwet's AC1 0.890), is therefore the test--retest consistency of one model annotation procedure, not inter-rater reliability, and the ``both annotators'' decision rule is close to a single test.
Correlated errors within and across model families~\citep{kohli2026ninejudges} and the gap between reproducible model annotation and construct validity~\citep{batzdorfer2026repro} both apply.
Frames are sampled, not continuous video.
None of the conclusions below is a statement about human judgement.

\paragraph{Results.}
\begin{itemize}
  \item \textbf{C3a (primary): inconclusive.} No predicate type showed a disagreement rate $\geq 10\%$ under both annotators, and the upper bounds of CheckGripperDistance and CheckGripperContact exceed 10\%, so the frozen rule yields neither ``holds'' nor ``does not hold''. IPW rates $P(\text{safe}\mid\text{predicate unsafe})$, annotators A/B: CheckDistance 0.02 [0.0, 0.065] / 0.04 [0.0, 0.104]; CheckGripperDistance 0.127 [0.033, 0.246] / 0.08 [0.007, 0.180]; CheckGripperContact 0.12 [0.005, 0.258] / 0.158 [0.040, 0.316]; Fall 0.041 [0.0, 0.104] for both. Truncating or removing IPW weights does not change the reading.
  \item \textbf{C3b (pipeline guard): passes.} On segments whose distance label flips when spurious zero distances returned by the simulator are corrected, both annotators' literal-rule answers sided with the corrected label in 10 of 13 definite answers (one-sided sign test $p=0.046$; 7 of 20 segments could not be judged), against 0 on control segments.
  \item \textbf{C3c (secondary): holds.} On segments where the two evaluator versions disagree, agreement with the annotators is higher for the current definitions: difference 0.652 [0.348, 0.885] (A) and 0.591 [0.273, 0.846] (B), both Holm-significant.
  \item \textbf{C3d (secondary): does not hold.} Disagreement is not concentrated in the pre-registered hard cases.
  \item Overall IPW agreement with the corrected predicate labels: 0.932 (A) and 0.928 (B).
\end{itemize}

\paragraph{Reading.}
The probe supports the annotation pipeline (C3b) and points in the direction of the current definitions on segments where the versions disagree (C3c), within the limits above.
It does not establish that any predicate measures unsafe behaviour (C3a is inconclusive), and the main text does not rely on it: the \texttt{Fall} defect is established by a policy-free probe and geometry, which do not involve annotators.

\section{The \pifast{} Entry}
\label{app:pi0fast}

\paragraph{What we observed in the main runs.}
We ran the released \pifast{} LoRA checkpoint (the checkpoint linked by its leaderboard entry, blob hashes matching the hub) through the official evaluator at both commits, with the official evaluation configuration.
Across all complete test jobs, 78,605 of 168,000 inference calls (0.468) failed to decode the generated FAST tokens into an action chunk (per-job range 0.436--0.537).
On failure the tokenizer substitutes an all-zero chunk, which de-normalizes to a fixed drifting action; the policy succeeded in 0 of 2,800 test episodes, each running to the 300-step limit.
Our wrapper calls the official evaluator and policy server and only adds timing, and our configuration matches the official evaluation file item by item; the other three entries showed no decoding errors in any live job.
Decoding failures of this kind are reported upstream for other FAST checkpoints~\citep{openpi_issue833}, at lower frequency.

\paragraph{Discriminating diagnostic (pre-registered, development split).}
To separate the checkpoint from the pipeline, we ran the LoRA checkpoint and the full-parameter (FFT) \pifast{} checkpoint from the same release through the identical pipeline on the same GPU type (H200), 10 development episodes each, with the decision rule fixed in advance (LoRA failure rate $\geq0.30$, FFT $\leq0.05$ with non-zero success $\Rightarrow$ checkpoint-specific).

\begin{table}[h]
\centering
\caption{\textbf{\pifast{} diagnostic.} Decode-failure rate over inference calls; 95\% CI clustered by episode (recomputed by the auditor from the order of log lines, assuming server and client log lines are written in order).}
\label{tab:w5}
\small
\begin{tabular}{@{}lllll@{}}
\toprule
Checkpoint & inference calls & decode failures & rate [95\% CI] & successes \\
\midrule
LoRA (released, leaderboard entry) & 600 & 262 & 0.437 [0.418, 0.460] & 0/10 \\
Full-parameter (same release) & 202 & 2 & 0.010 [0.000, 0.022] & 10/10 \\
\bottomrule
\end{tabular}
\end{table}

The reading is \emph{checkpoint-specific}.
In the same pipeline the full-parameter checkpoint runs normally, which rules out the shared FAST decoding path, tokenizer and software environment as the cause; the failure belongs to the released LoRA checkpoint or its released configuration.
The two checkpoints use different training configurations, so the diagnostic does not say which of the two is at fault.
Because decoding is greedy, the two trials per task are identical, so each checkpoint produced 5 distinct rollouts.
A reference run of the LoRA checkpoint on a Blackwell GPU before the diagnostic was frozen gave a similar failure rate (0.517) and is not part of the reading.

\paragraph{How we report it.}
We exclude the LoRA entry as technically invalid and do not use it as evidence that the leaderboard is irreproducible; this development-scale diagnostic makes no claim about how the leaderboard's \pifast{} numbers were obtained.
Excluding it reduces the significant reversals of the exploratory test from 8 to 3 and leaves both counted reversals unchanged.

\end{document}